\documentclass[letterpaper, 10 pt, conference]{ieeeconf}  

\IEEEoverridecommandlockouts                              

\usepackage{times}
\usepackage{latexsym}
\usepackage{amsmath}
\usepackage{comment}
\usepackage{booktabs}
\usepackage{algorithm} 
\usepackage{algpseudocode} 
\usepackage{amssymb}
\usepackage{longtable}

\usepackage[most]{tcolorbox}
\usepackage{xcolor}

\usepackage{wrapfig}

\usepackage{listings}
\usepackage{xcolor}
\usepackage{courier}
\usepackage{multirow}
\usepackage[hidelinks]{hyperref}

\title{\LARGE \bf
LookPlanGraph: Embodied Instruction Following Method with \\VLM Graph Augmentation
}

\author{Anatoly O. Onishchenko$^{1}$, Alexey K. Kovalev$^{2,1}$ and Aleksandr I. Panov$^{2,1}$
\thanks{$^{1}$MIRAI, Moscow, Russia
        {\tt\small (\{onishchenko.a, kovalev.a\}@miriai.org)}}
\thanks{$^{2}$Cognitive AI Systems Lab, Moscow, Russia
         {\tt\small (panov@cogailab.com)}
}
}

\begin{document}

\maketitle

\begin{abstract}
Methods that use Large Language Models (LLM) as planners for embodied instruction following tasks have become widespread. To successfully complete tasks, the LLM must be grounded in the environment in which the robot operates. One solution is to use a scene graph that contains all the necessary information. 
Modern methods rely on prebuilt scene graphs and assume that all task-relevant information is available at the start of planning. However, these approaches do not account for changes in the environment that may occur between the graph’s construction and the task execution.
We propose LookPlanGraph -- a method that leverages a scene graph composed of static assets and object priors.
During plan execution, LookPlanGraph continuously updates the graph with relevant objects, either by verifying existing priors or discovering new entities.
This is achieved by processing the agent’s egocentric camera view using a Vision Language Model. 
We conducted experiments with changed object positions VirtualHome and OmniGibson simulated environments, demonstrating that LookPlanGraph outperforms methods based on predefined static scene graphs. To demonstrate the practical applicability of our approach, we also conducted experiments in a real-world setting.
Additionally, we introduce the GraSIF (\underline{Gra}ph \underline{S}cenes for \underline{I}nstruction \underline{F}ollowing) dataset with automated validation framework, comprising 514 tasks drawn from SayPlan Office, BEHAVIOR-1K, and VirtualHome RobotHow. Project page available at \url{https://lookplangraph.github.io/}.

\end{abstract}

\section{Introduction}

The pursuit of embodied agents, such as robots, that can understand and execute complex human instructions in dynamic environments is a fundamental goal of Embodied Artificial Intelligence. 
Recent advances in large language models have shown promising capabilities in natural language reasoning and planning~\cite{ahn2022can,kovalev2022application,sarkisyan2023evaluation,singh2023progprompt}. 
However, for effective task execution, these models must be grounded in the physical environment -- often achieved through scene graphs that capture objects and their relationships~\cite{gu2023conceptgraphsopenvocabulary3dscene}.

Current methods often~\cite{rana2023sayplan,liu2023llmpempoweringlargelanguage,liu2024deltadecomposedefficientlongterm} assume a fully known environment and generate complete action sequences in a single pass. This approach significantly limits their effectiveness in dynamic environments~\cite{pchelintsev2025lera}, where objects can appear or change position over time (see Figure~\ref{fig:dynamic_env}).
While several methods for dynamically building graphs have been proposed~\cite{honerkamp2024language,tang2025openin}, these approaches primarily focus on vision language navigation and do not fully address manipulation challenges, such as inferring goal states from complex instructions and generating feasible plans for multiple object rearrangement.

\begin{figure}[t]
  \includegraphics[width=\columnwidth]{images/problem.png}
  \caption{Static planners rely solely on predefined scene graphs, making them ineffective when objects are missing or misplaced.  In contrast, dynamic planners can explore the environment in real time and update the scene graph to account for newly discovered objects. This allows them to successfully execute tasks.}
  \label{fig:dynamic_env}
  \vspace{-15pt}
\end{figure}

To address these limitations, we propose LookPlanGraph, a method that integrates graph-based planning with dynamic scene updates.
At the core of our approach is the \textbf{Memory Graph}, which consists of three key components: 1) the general scene structure; 2) the immovable objects (assets) previously interacted with; and 3) the probable positions of the movable objects (priors). Complementing the memory graph is the \textbf{Scene Graph Simulator} (SGS), which validate and refines actions generated by the language model (LM) and updates the Memory Graph to reflect environmental changes, enabling the LM to dynamically choose actions based on the current state of environment.
This design enables the integration of a \textbf{graph augmentation module}, which can discover new objects or update their positions in a changed environment. The agent explores the environment using available assets, capturing images from predefined poses. These images are then processed by a Vision Language Model (VLM) to infer object relationships and properties, incorporating the detected objects into the accurate scene representation.

We comprehensively evaluate LookPlanGraph in both dynamic and static scenarios, including real-world office experiments to showcase its practical applicability. Additionally, we assess the performance of the graph augmentation module and perform an ablation study on the memory and simulation components. Given the lack of graph-based instruction-following datasets, we introduce GraSIF (\underline{Gra}ph \underline{S}cenes for \underline{I}nstruction \underline{F}ollowing), a dataset constructed with manipulation tasks from SayPlan Office~\cite{rana2023sayplan}, Behaviour-1k~\cite{li2024behavior}, and RobotHow~\cite{puig2018virtualhome}.
Our results show that LookPlanGraph achieves competitive planning performance in static environments and outperforms existing planning methods in dynamic settings.

In summary, our main contributions are:
\begin{enumerate}
    \item \textbf{LookPlanGraph:} We introduce a method for adaptive and context-aware planning that achieves the highest performance in dynamic environments. 
    \item \textbf{Graph augmentation module:} We propose a module that utilizes a VLM to dynamically update scene graphs based on real-time observations.
    \item \textbf{GraSIF Dataset:} We present the benchmark of 514 tasks, along with a lightweight, automated validation framework, designed to evaluate graph-based instruction following in household manipulation scenarios.
\end{enumerate}

\section{Related Works}
\subsection{Embodied Planning}
Embodied instruction following involves generating sequences of actions to accomplish goals described in natural language within a given environment~\cite{kachaev2025mind}. Traditional methods follow two main approaches. \textit{Symbolic planning}, where entities, actions, and goals are represented as discrete symbols, such as in the Planning Domain Definition Language~\cite{agia2022taskography} (PDDL) or Temporal Logic~\cite{doherty2001talplanner}. 
\textit{Hierarchical policy learning} for grounding language instructions into structured sequences of actions~\cite{ahn2022can, zhang2022danli, zheng2022jarvis}.
While these methods are effective in controlled environments, they struggle with scalability and generalization. Adapting them to new tasks often requires additional analysis and training or the construction of new symbolic representations to account for changes in task instructions.

Recently, LM have gained traction in task planning~\cite{huang2022language,huang2022inner,yao2023react,rana2023sayplan,grigorev2024common,grigorev2025verifyllm,chen2025schema} due to their strong in-context learning and reasoning abilities.
Some approaches \cite{liu2023llmpempoweringlargelanguage, liu2024deltadecomposedefficientlongterm, dai2024optimal}, combine LMs with symbolic planners by having the model generate PDDL descriptions.
A major challenge in LM-based planning is grounding in the environment. Certain methods work in textual~\cite{yao2023react, wang2022scienceworld} or simulated settings~\cite{song2023llm, huang2022inner, booker2024embodiedrag}, where the environment itself provides feedback or additional context to help guide the model’s output, support that may not be available in real-world scenarios.
SayPlan~\cite{rana2023sayplan} tackles this issue by constructing feedback from pre-constructed 3D scene graph. While this enhances plan executability, it assumes that the environment remains static after the map is created. This assumption limits the method’s applicability in dynamic real-world settings.
Our approach allows to incorporate environmental information during planning and adjusts LM chosen actions using the graph structure.

\subsection{Dynamic Instruction Following}
Several approaches~\cite{honerkamp2024language, tang2025openin} dynamically build graphs guided by LMs. Yet, they target vision-language navigation tasks aimed mainly at locating objects and do not address tasks requiring goal state comprehension and rearrangement of multiple objects. For example, in MoMa-LM~\cite{honerkamp2024language} manipulation is limited to open and close actions, designed to address occlusion caused by closed assets and don't involve object relocation. ``Success'' is defined as observing the target object and calling ``done()''.

VeriGraph~\cite{ekpo2024verigraph} uses a VLM to construct task-specific graph representations and generate plans accordingly. While VLMs show strong potential for graph construction, VeriGraph has only been evaluated in tabletop environments, limiting its applicability to more complex scenarios.
Our method extends this idea to multi-room environments, enabling a broader range of tasks using only image-based observations.

\section{Problem Formulation}
We tackle the problem of enabling an autonomous mobile manipulator to strategically plan, traverse, and interact with objects in domestic environments. Object locations may vary between initial environment mapping and actual task performance. The goal is to synthesize action sequences that conform to a given natural language instruction $I$.

To achieve this, we make two key assumptions:
We assume a pre-constructed graph representation of the environment, denoted as $G = (V, E)$, where $V$ and $E$ represent the sets of vertices and edges, respectively. The graph $G$ is structured hierarchically into four primary levels: scene, places, assets, and objects. The scene is represented at the top level by a single \textit{scene} node that connects to all place nodes. Each place contains nodes for immovable assets, which in turn hold nodes for movable objects. These object nodes can be linked to their parent assets or to other objects. Both asset and object nodes have defined states and affordances. The edges $E$ capture ``inside'' spatial relationship within higher-level nodes. For object-object and object-asset connections, an additional type ``ontop'' present. Graphs of this form can be constructed using existing techniques ~\cite{hughes2022hydrarealtimespatialperception,gu2023conceptgraphsopenvocabulary3dscene,tang2025openin}.
It is important to note that LookPlanGraph can function without initial information about objects, but this information is still included as it provides valuable priors for guiding the search process, potentially derived from previous task executions or the initial graph construction.
We also assume that the robot has access to high-level manipulation actions $\mathbf{A}$, for which approaches from prior works~\cite{wen2025tinyvla} can be used.

Our objective is to create high-level decision-making policy denoted by $\pi(\mathbf{A} \mid I, G)$ that is capable of generating an action $\mathbf{A}$ that is executable within the environment. This policy should enable the embodied agent to follow an natural language instruction $I$ 
by appropriately modifying the environment's state.

\section{Method}
\subsection{Overview}

\begin{figure*}[t]
\centering
\includegraphics[width=0.85\textwidth]{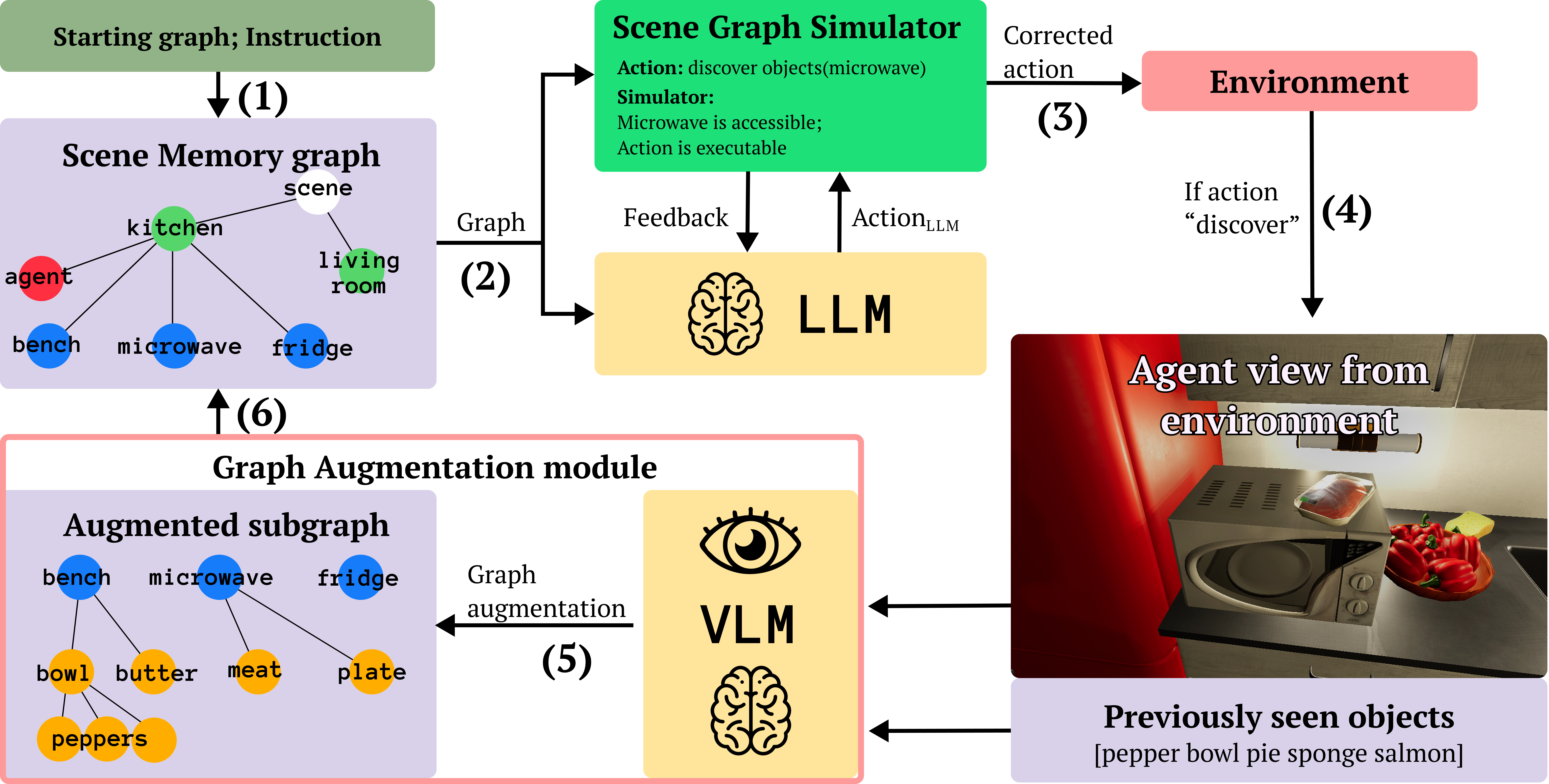}
  \caption{\textbf{LookPlanGraph Overview:} The LookPlanGraph starts with an instruction and a static environment graph~\textbf{(1)}. A scene memory graph, initially a copy of the starting graph, is processed by the LM with the task description and is also sent to the Scene Graph Simulator~\textbf{(2)}. The LM suggests an action, which the Simulator checks for feasibility. If feasible, the action changes the environment and updates the SMG~\textbf{(3)}. For actions requiring visual feedback (e.g., ``discover\_objects''), the environment sends an egocentric camera view to the VLM~\textbf{(4)}. The VLM processes this image, along with previously seen objects from the SMG, to generate an augmented subgraph~\textbf{(5)}, which updates the SMG~\textbf{(6)}. This cycle~\textbf{(2-6)} repeats until the LM decides that the task is complete.}
  \label{fig:LookPlanGraph}
  \vspace{-10pt}
\end{figure*}

We introduce LookPlanGraph -- a method that dynamically generates and refines actions using a graph representation of the scene. As illustrated in Figure~\ref{fig:LookPlanGraph}, the process begins by copying the initial graph to a memory graph, with all object nodes initially marked as unseen.
The method then proceeds with the iterative process outlined in Algorithm~\ref{alg:lookplangraph} (steps 4-14), where it repeatedly generates and executes actions in the environment.

\begin{algorithm}
    \caption{LookPlanGraph}
    \begin{algorithmic}[1]
        \Require Instruction $I$, Starting graph $G$
        \State \textbf{Modules:} LM -- Language Model, GAM -- Graph Augmentation Module, ENV -- Environment, SGS -- Scene Graph Simulator
        \State \textbf{Variables:} $M$ -- Scene Memory graph; $F$ -- feedback string; $\text{action}_{\text{lm}}$ -- action chosen by LM; $\text{action}_{\text{corr}}$ -- action corrected by SGS.

        \State \textbf{Initialize:} $M \gets G$
        \Repeat
            \Repeat
                \State $\text{action}_{lm} \gets$ LM$(M, I, F)$
                \State $\text{action}_{corr}, F$ $\gets$ SGS$(M, \text{action}_{lm})$
            \Until{$F == None$}
            \State \textit{ENV}($\text{action}_{corr}$)
            \If{$\text{action}_{corr} ==$ ``discover\_objects''}
                \State new\_nodes $\gets$ GAM$(\text{ENV}, M)$
                \State $M$.append(new\_nodes)
            \EndIf
        \Until{$\text{action}_{corr} ==$ ``done''}
    \end{algorithmic}
    \label{alg:lookplangraph}
\end{algorithm}

\textbf{LM Decision-Making (6):} 
The LM is assigned the role of an agent and provided with a set of possible actions (\textit{discover objects, go to, pick up, rearrange, open, close, turn on, turn off, done}). It receives the few-shot examples of input-output interactions, a text-serialized memory graph, instruction, previous reasons and actions, and, if available, feedback. An example prompt structure is shown in Figure~\ref{fig:lpg_prompts}. 
Based on this input, the LM generates a text description detailing the reasoning behind the decision and the subsequent action to be executed.
    
\textbf{Simulation and Feedback (5-8):} 
The output from the LM is sent to the SGS to verify the action's feasibility. The SGS, referencing the memory graph, either corrects the action to fit the current environment or returns it to the LM for replanning in case of hallucinations or structural errors. When the action is deemed valid, the SGS updates memory graph to reflect the changes in the environment.

\textbf{Environment Interaction (9):} 
Once validated by the simulator, the action is executed in the environment. For exploratory actions like \textit{``discover\_objects''}, the Graph Augmentation Module is triggered to expand the memory graph.

\textbf{Graph Augmentation via VLM (10-13):}
Upon activation, the VLM processes image from agent egocentric camera, identifying objects and adding their corresponding nodes to the memory graph.

This loop continues until the LM selects the ``done'' action, a sequence of three infeasible actions, or upon reaching a predefined action limit. Notably, the same VLM employed for visual graph augmentation can also serve as the LM for action decision-making, enhancing operational efficiency and reducing system complexity.

\subsection{Memory Graph}
Real-world environments are inherently dynamic, meaning that object positions in a pre-constructed graph may not remain accurate during execution. To account for this, all objects in the initial graph are marked as unseen at the start of the planning process, indicating that their presence and location must be confirmed through perception.

Building a plan dynamically requires prompting the scene graph at each decision point. However, querying the entire graph for every action would incur a high interaction cost and potentially degrade the model's reasoning performance due to long prompt lengths.
To mitigate this, a memory graph is used as an intermediate representation, filtering and abstracting only the most relevant information. This graph includes a list of known places, the agent node, and only those assets and unseen objects within the agent's immediate surroundings (e.g., the same room).

\begin{figure}[t]
  \includegraphics[width=\columnwidth]{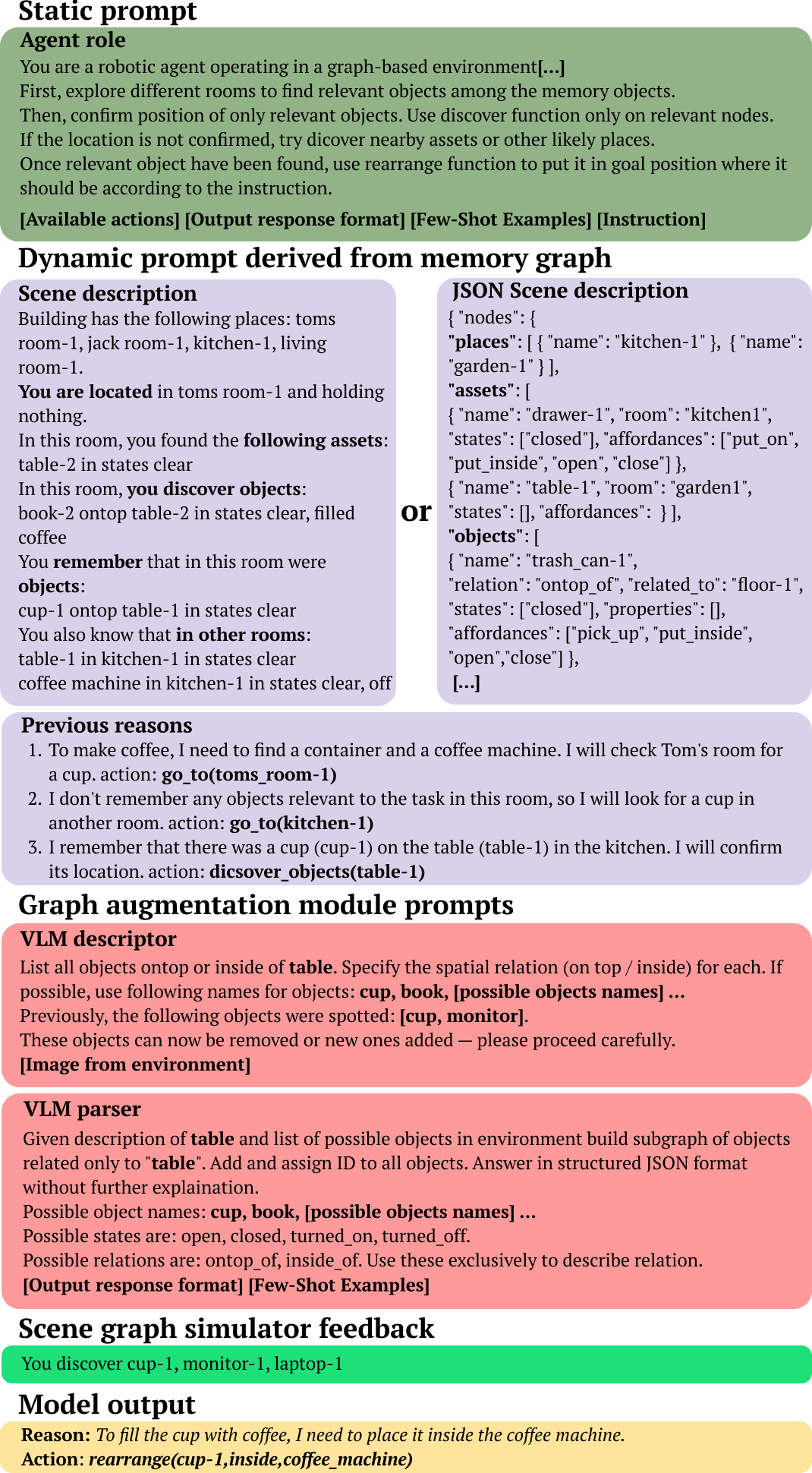}
  \vspace{-15pt}
  \caption{LookPlanGraph Prompt Structure. The prompt consists of a static section that remains unchanged across all instructions, a dynamic section constructed from the current memory graph and previous actions, and a VLM-specific prompt used for graph augmentation.}
  \vspace{-20pt}
  \label{fig:lpg_prompts}
\end{figure}

To maintain continuity and accumulate knowledge during task execution, the memory graph is incrementally updated with all discovered objects and interacted-with assets. These elements remain visible even if the agent moves to a different location, allowing the graph to serve as a persistent memory module, supporting the agent in recalling and utilizing previously acquired information. When the graph augmentation module is used, unseen objects are removed or replaced with objects extracted from the environment.
See Figure~\ref{fig:lpg_prompts} for text serialized memory graph representation.

\subsection{Scene Graph Simulator}
The SGS provides a dynamic and maintained representation of the environment, ensuring that the agent remains contextually grounded during task execution. It corrects erroneous LM actions using two primary mechanisms: re-prompting and rule-based correction.
Errors typically fall into two categories: 1) hallucinations, which occur when the LM references non-existent objects or actions, and response structure errors; and 2) execution errors, which occur when the agent attempts infeasible actions in the current environment, such as picking up an item from a closed container.

To address hallucinations, the SGS uses a reprompting strategy similar to SayPlan~\cite{rana2023sayplan}, providing the LM with corrective feedback. For execution errors, it applies rule-based adjustments to reduce costly reprompting. For example, if the LM suggests \textit{pick up(apple-1)} while the apple is in a closed container, the SGS can automatically add \textit{open(container)} as a prerequisite.
We extend this approach by introducing a higher-level ``rearrange'' action for the LM, allowing the SGS to autonomously generate intermediate steps without requiring the explicit generation of each individual action.

\subsection{Graph Augmentation Module}
For visual grounding within the environment, the LM can use the ``discover\_objects'' action to inspect assets or objects, adding identified objects to the memory graph. An example of this interaction is shown in the lower part of Figure~\ref{fig:LookPlanGraph}. The procedure starts by capturing an image from the agent’s current field of view. This image, along with a list of known object names in the scene, and prior information about objects is passed to VLM. The VLM is instructed to describe the scene and detect objects, their states, and their spatial relationships to the observed asset. This description then parsed into a structured format. The identified nodes are incorporated into the memory, making them accessible for subsequent interactions. The prompts for the VLM is provided in Figure~\ref{fig:lpg_prompts}.

\section{The GraSIF Dataset}
Planners that utilize graph scene representations are gaining popularity in the research community. However, the lack of publicly available task-specific scene graph datasets, or reliance on computationally intensive physics simulators, complicates the evaluation of planning methods. As a result, many researchers resort to proprietary data, hindering consistent comparison across approaches.
To address this gap, we introduce GraSIF (\underline{Gra}ph \underline{S}cenes for \underline{I}nstruction \underline{F}ollowing) -- a benchmark comprising 514 instruction-following tasks represented as scene graphs in household environments, paired with a lightweight automated validation framework.
GraSIF integrates environmental data from three sources: BEHAVIOR-1K~\cite{li2024behavior}, VirtualHome RobotHow~\cite{puig2018virtualhome}, and SayPlan Office~\cite{rana2023sayplan}. 
It focuses specifically on \textit{mobile manipulation} tasks, where an agent must find a sequence of manipulation actions to complete a given instruction. For each environment, GraSIF provides unambiguous natural language instructions along with both the initial and goal scene graphs, representing the state before and after task execution. All graphs are constructed using consistent filtering rules to ensure comparability across tasks. For details on the graph construction and filtering process, see Appendix~\ref{app:datset_prep}.

To evaluate action sequences generated by planners, we first identify task-relevant nodes whose position or state changes between the initial and goal states. We then run the planner's sequence in a scene graph simulator and compare the resulting goal state to the expected state of these changed nodes.
Since some tasks can have unambiguous but multiple valid solutions (e.g., bringing any three bottles to a table), we validate by checking the required number of relations for key objects rather than exact IDs. The precision of the plan is calculated by comparing the number of correctly placed object nodes and assets states to the expected in the goal state. A plan is considered successful if all objects are in the correct positions, assets are in the desired states, and the plan does not contain any infeasible actions.
This approach enables fast automated validation without relying on complex physics simulations.

In total, the GraSIF dataset comprises 10 diverse scenes and 514 tasks, each emphasizing different aspects of embodied planning.
The largest environment, SayPlan Office, spans 37 rooms, providing a complex and extensive planning space.
In contrast, BEHAVIOR-1K tasks, while typically set in smaller, one-to-two-room environments, feature long-horizon tasks that can require up to 40 actions to complete.
The VirtualHome environment consists of 4 rooms and includes 195 object nodes, resulting in a densely populated setting.

\begin{table}[t]
    \centering
    \caption{The GraSIF dataset statistics.}
    \begin{tabular}{llllll}
        \toprule
        \textbf{Subdata} & \textbf{Tasks} & \textbf{Rooms} & \textbf{Nodes}  & \textbf{Actions} \\
        \midrule
        SayPlan Office      & 29  &  \textbf{37}  & \textbf{202.6}   &  4.2 \\
        BEHAVIOR-1K        & 177 &  1.23   & 12.1    &  \textbf{9.8} \\
        VirtualHome         & \textbf{308} &  4   & 195.7   &  3.1 \\
        \bottomrule
    \end{tabular}
    \vspace{-10pt}
    \label{tab:datasets}
\end{table}
The characteristics of the integrated datasets are summarized in Table~\ref{tab:datasets}, where the values for ``Rooms'', ``Nodes'', and ``Actions'' represent the mean across all tasks.

\section{Experiments}
\subsection{Baselines}
We evaluate LookPlanGraph against five baseline methods that incorporate LM as graph planners. 
\textbf{LLM-as-P}~\cite{liu2023llmpempoweringlargelanguage} produces a complete sequence of actions by utilizing the task description, the available actions within the environment, and a one-shot example.
\textbf{LLM+P}~\cite{liu2023llmpempoweringlargelanguage} translates natural language task descriptions into PDDL problem formulations and then relies on a classical planner to generate plans. In practice, this approach struggles with syntactic and semantic errors in LLM-generated PDDL, leading to parsing failures. To overcome this, instead of using the original Manipulation domain, we designed a domain specifically tailored to our graph-based environment, enabling evaluation of instruction-following tasks.
SayPlan~\cite{rana2023sayplan} operates in two stages: first, it extracts a task-relevant subgraph using LM-based semantic search, and then it generates a full plan over this subgraph, iteratively refining it based on scene graph feedback. Due to the lack of open-source code, we created our own implementation for this study.
However, observing frequent stage confusion in smaller language models, we developed \textbf{SayPlan Lite}. This variant enforces a strict separation between its search and planning phases: only the search API is available during the search phase, and only the planning API during the planning phase. Furthermore, the action set is simplified by replacing the original \textit{access} and \textit{release} operations with the more direct \textit{put on} and \textit{put inside} actions, which helps to unify the low-level action set across all methods.
\textbf{ReAct}~\cite{yao2023react} is a dynamic approach that uses feedback from the simulation as exploration and saves full interactions history during planning. To incorporate ReAct into graph planning, we adapt the SayPlan simulator to provide feedback on each action in the form of an ALFWorld~\cite{shridhar2020alfworld} textual environment, as it was implemented in the original paper. For tasks requiring perception, we employ our graph augmentation module. Appendices~\ref{app:baselines} and~\ref{app:prompts} provide further details on the baseline implementations, including prompt structures and domain specifications.

\subsection{Experimental setup}

To assess LookPlanGraph’s \textbf{performance in dynamic environments} we conducted experiments using VirtualHome and OmniGibson simulators.
In VirtualHome, we designed 50 tasks across three categories: placing an object in a refrigerator, heating an object in a microwave, and washing dishes in a dishwasher. 
OmniGibson does not provide native support for high-level actions; therefore, a scene graph simulator was utilized to emulate these actions within the environment. For perception actions, we collected images from simulations for each asset containing the required objects, using 50 tasks from the Behaviour-1K dataset. To introduce dynamic changes, we modified the positions of objects in the initial scene graph for half of the tasks in both environments.

\textbf{Graph augmentation capabilities} were evaluated using the images gathered during simulation. We measured accuracy with the F1-score by comparing the nodes and edges of the generated scene graphs against ground truth. A perfect match of all nodes and edges with real graph was considered a successful generation.

To assess solely \textbf{planning performance} across a broader range of tasks, we conducted experiments on the GraSIF dataset. Since GraSIF contains only graph data, the augmentation module was modified to include nodes connected to the explored nodes. Tokens for graph augmentation were excluded from the TPA calculation, to compare cost of planing exclusively.
For evaluation, we used \texttt{gpt-4o-2024-08-06}~\cite{achiam2023gpt} and Llama3.2-90b-vision~\cite{dubey2024llama} as VLM,  Llama3.3 and Gemma3~\cite{team2024gemma} as LM.

Experiments involving GPT-4o are conducted via the OpenAI API.
Open source models are executed on two Tesla V100 GPUs, each with 32GB of VRAM. Simulations in VirtualHome and BEHAVIOR-1K are performed with an Nvidia RTX 2080 GPU.

\subsection{Metrics}
\label{sec:metrics}
We use the following metrics for evaluation. 
\textbf{Success Rate} (\textbf{SR}) is the ratio of successfully completed tasks to all tasks. Task is considered successfully completed if all graph nodes are correctly transformed to match their goal configuration.
\textbf{Average Plan Precision} (\textbf{APP}) is the ratio of correctly modified nodes in the generated plan relative to the total number of changed nodes. APP measures the precision of the method in modifying graph nodes to achieve the goal state. 
\textbf{Tokens Per Action} (\textbf{TPA}) measures the computational efficiency and cost of a method. It is calculated as the ratio of the total number of tokens used during planning to the number of actions generated.
Additional details on used metrics provided in Appendix~\ref{app:metrics_eq}.

\section{Results}
\subsection{Performance in Dynamic Environment}

The results of our evaluation in dynamic environments are summarized in Table~\ref{tab:sim_results}. LookPlanGraph consistently outperforms all baselines in both SR and APP by effectively leveraging visual descriptions and adapting between graph construction and execution. ReAct underperforms due to an overreliance on previous actions and insufficient use of semantic context, often repeating similar searches (e.g., across multiple drawers). In contrast, LookPlanGraph integrates scene descriptions and object priors at every step, enabling more targeted searches and improving SR.

\begin{table}[b]
\vspace{-10pt}
\centering
\caption{Methods performance in simulated dynamic environments.}
\setlength{\tabcolsep}{3pt}
\begin{tabular}{p{2.1cm}cc|cc|cc|cc}
\toprule
\multirow{2}{*}{\textbf{Method}} & \multicolumn{4}{c}{\textbf{VirtualHome}} & \multicolumn{4}{c}{\textbf{OmniGibson}} \\
& \multicolumn{2}{c}{\textbf{Llama3.2}} & \multicolumn{2}{c}{\textbf{GPT-4o}} & \multicolumn{2}{c}{\textbf{Llama3.2}} & \multicolumn{2}{c}{\textbf{GPT-4o}} \\
\cmidrule(lr){2-3} \cmidrule(lr){4-5} \cmidrule(lr){6-7} \cmidrule(lr){8-9}
& SR$\uparrow$ & APP$\uparrow$ & SR$\uparrow$ & APP$\uparrow$ & SR$\uparrow$ & APP$\uparrow$ & SR$\uparrow$ & APP$\uparrow$ \\
\midrule
LLM-as-P        & 0.16 & 0.53 & 0.44 & 0.65 & 0.16 & 0.36 & 0.33 & 0.59 \\
LLM+P           & 0.00 & 0.38 & 0.32 & 0.58 & 0.21 & 0.40 & 0.37 & 0.62 \\
SayPlan         & 0.21 & 0.53 & 0.38 & 0.59 & 0.10 & 0.26 & 0.12 & 0.31 \\
SayPlan Lite    & 0.39 & 0.65 & 0.48 & 0.67 & 0.27 & 0.55 & 0.40 & 0.67 \\
ReAct           & 0.30 & 0.64 & 0.22 & 0.50 & 0.33 & 0.52 & 0.41 & 0.64 \\
LookPlanGraph   & \textbf{0.60} & \textbf{0.68} & \textbf{0.52} & \textbf{0.63} & \textbf{0.35} & \textbf{0.60} & \textbf{0.42} & \textbf{0.69} \\
\midrule
ReAct$_{GT}$    & 0.50 & 0.74 & 0.34 & 0.58 & 0.63 & 0.75 & 0.66 & 0.78 \\
LookPlanGraph$_{GT}$ & \textbf{0.92} & \textbf{0.96} & \textbf{0.86} & \textbf{0.92} & \textbf{0.74} & \textbf{0.86} & \textbf{0.71} & \textbf{0.84} \\
\bottomrule
\end{tabular}
\label{tab:sim_results}
\end{table}

We also observe notable differences between LlamA3.2-90B and GPT-4o. LlamA3.2-90B attains higher SR due to more reliable search behavior, whereas GPT-4o, despite its stronger augmentation capability, performs worse in search by overemphasizing past actions. Static planners fail to surpass 50\% SR, revealing fundamental limitations in dynamic scenarios. 
With ground-truth augmentation, the performance gap between LookPlanGraph and other planners widens further, underscoring the effectiveness of our planning framework.

The LLM+P pipeline with LLama3.2-90B exhibits significant challenges with predicate misalignment. Frequent errors described in Appendix~\ref{app:failurecases} include substituting valid predicates with undefined ones (e.g., \texttt{on-state} replaced by \texttt{on}) and mishandling negations (e.g., \texttt{(not (on-state))} replaced by \texttt{off}). While increasing the model size has partially alleviated these issues, they remain a persistent problem.

\subsection{Graph Augmentation Capability} 
Table~\ref{tab:mainresults2} presents the results of the graph augmentation module. Using GPT-4o, accurate graphs were generated in 60\% of cases across various assets in house and store environments within OmniGibson. Similar performance was observed in VirtualHome, though accuracy decreased with increasing object count: the VLM often underestimates repeated instances (e.g., adding 3 apples instead of the 8 present in the image).
Reducing the size of the model resulted in an overall decrease in recognition capabilities, coupled with problems in producing structured output, reducing the SR to 33\%.

\begin{table}[t]
    \caption{Graph augmentation capability results.}
    \centering
    \begin{tabular}{lccc|ccc}
    \toprule
    & \multicolumn{3}{c}{\textbf{VirtualHome}} & \multicolumn{3}{c}{\textbf{OmniGibson}} \\
    \midrule
    \textbf{Model} & \textbf{F1$_N$} & \textbf{F1$_E$} & \textbf{Success} & \textbf{F1$_N$} & \textbf{F1$_E$} & \textbf{Success} \\
    \midrule
    Llama3.2-90b    & 0.78 & 0.73 & 0.26 & 0.67 & 0.67 & 0.33 \\
    GPT-4o          & \textbf{0.87} & \textbf{0.84} & \textbf{0.44} & \textbf{0.85} & \textbf{0.88} & \textbf{0.59} \\
    \bottomrule
    \end{tabular}
    \label{tab:mainresults2}
    \vspace{-10pt}
\end{table}

\subsection{Planning Performance}

\begin{table}[b]
     \vspace{-10pt}
     \caption{Methods comparison on the GraSIF dataset.}
     \centering
     \setlength{\tabcolsep}{3pt}
     \begin{tabular}{p{1.8cm}c@{\hskip 3pt}c@{\hskip 3pt}c|c@{\hskip 3pt}c@{\hskip 3pt}c|c@{\hskip 3pt}c@{\hskip 3pt}c}
     \toprule
     & \multicolumn{3}{c}{\textbf{SayPlan Office}}
     & \multicolumn{3}{c}{\textbf{BEHAVIOR-1K}}
     & \multicolumn{3}{c}{\textbf{RobotHow}}\\ 
     \cmidrule(lr){2-4} \cmidrule(lr){5-7} \cmidrule(lr){8-10}
     \textbf{Method} 
     & SR$\uparrow$ & APP$\uparrow$ & TPA$\downarrow$ 
     & SR$\uparrow$ & APP$\uparrow$ & TPA$\downarrow$ 
     & SR$\uparrow$ & APP$\uparrow$ & TPA$\downarrow$ \\
     \midrule 
     LLM-as-P        & 0.47 & 0.59 & 1409
                     & 0.39 & 0.53 & \textbf{178} 
                     & 0.44 & 0.51 & 3417 \\
     LLM+P           & 0.07 & 0.21 & 1945 
                     & 0.33 & 0.37 & 160 
                     & 0.30 & 0.38 & 5396 \\
     SayPlan         & 0.46 & 0.59 & 3697 
                     & 0.36 & 0.43 & 1888 
                     & 0.86 & 0.87 & 5576 \\
     SayPlan Lite    & 0.53 & 0.68 & \textbf{1368}
                     & \textbf{0.61} & 0.76 & 524 
                     & 0.84 & 0.89 & 4641 \\
     ReAct           & 0.38 & 0.64 & 2503 
                     & 0.47 & 0.61 & 1713 
                     & \textbf{0.89} & \textbf{0.91} & \textbf{1322} \\
     LookPlanGraph   & \textbf{0.62} & \textbf{0.73} & 1989 
                     & 0.60 & \textbf{0.77} & 1472 
                     & 0.87 & 0.89 & 2653 \\
     \bottomrule
     \end{tabular}
     \label{tab:mainresults1}
\end{table}

Table~\ref{tab:mainresults1} presents the performance of various planning methods across three planning domains in the GraSIF dataset: SayPlan Office, BEHAVIOR-1K, and RobotHow.
In the SayPlan Office domain dynamic ReAct approach struggles to recognize correctnes of environment state only from feedback provided on actions frequently making wrong assumtion on task completness. LookPlanGraph surpass this limitation demonstrates the highest SR and APP, even outperforming methods that rely on complete scene graph.

In Behaviour-1k, as the number of changed objects in a task increase, the performance gap between SayPlan Lite and LookPlanGraph narrows as LM struggles to retrieve information about multiple objects states at once, stopping when only part of task executed.
Unlike SayPlan Lite, the standard SayPlan approach require access actions, which confuses LM and often resulting in early termination due to feedback interactions limit.

In the RobotHow benchmark, methods using the full scene graph for planning struggled to consistently retrieve key nodes. ReAct, rely on feedback as scene representation, was more effective at accessing relevant information and more token-efficient in dence populated scene. However, it offered only a marginal improvement in 2\% over LookPlanGraph, and show limited generalization in more complex environments. 
These results, along with the observation that different methods outperformed others on TPA metrics depending on the environment, highlight the necessity of evaluation across diverse environments, a capability supported by GraSIF.

\subsection{Ablation study}

\begin{table}[t]
  \centering
  \caption{Ablation study results on the SayPlanOffice dataset.}
  \begin{tabular}{lccc}
    \toprule
    \textbf{Method / Ablation} & \textbf{SR $\uparrow$} & \textbf{APP $\uparrow$} & \textbf{TPA $\downarrow$} \\
    \midrule
    LookPlanGraph           & 0.62 & 0.73 & 1989 \\
    Json graph              & 0.55 & 0.67 & 2599 \\
    Full Graph              & 0.65 & 0.78 & 4044 \\
    No Priors               & 0.51 & 0.68 & 1767 \\
    No Correction           & 0.24 & 0.40 & 1682 \\
    \midrule
    GPT-4o                  & 0.76 & 0.83 & 1752 \\
    Llama3.2-90b            & 0.59 & 0.72 & 2004 \\
    Gemma3-12b              & 0.31 & 0.45 & 2095 \\
    \bottomrule
  \end{tabular}
  \label{tab:ablation}
  \vspace{-10pt}
\end{table}

We analyze key components of LookPlanGraph through ablation studies examining graph representation, system components, and model scalability.

\textbf{Graph Representation.} JSON-based serialization reduces SR and APP while increasing token usage by 30\%, showing that natural language descriptions are more effective than structured formats due to absence of explicit format symbols and repeating properties names. Providing to LM full scene graph improves performance but doubles token cost, demonstrating memory graph's efficiency-performance balance.

\textbf{Component Analysis.} Removing object priors slightly reduces SR, confirming that while LookPlanGraph functions without prior knowledge, leveraging available information enhances search effectiveness. The action correction mechanism proves critical, disabling it causes dramatic performance drops, as minor planning errors introduce infeasible actions into resulting plans.

\textbf{Model Scalability.} Performance scales with model capability: GPT-4o achieves highest performance while maintaining efficiency, Llama3.2-90b delivers competitive results with slightly higher token usage, and smaller models like Gemma3-12b show significant degradation but remain functional, which demonstrates LookPlanGraph's adaptability across model sizes.

\subsection{Real-world evaluation}

To validate LookPlanGraph's practical applicability, we conducted experiments in a real office environment containing 7 rooms, 75 assets, and 130 objects. We manually constructed a scene graph representing this environment and designed 15 instruction-following tasks targeting commonly moved items such as cups, keys, clothing, and office supplies that require goal understanding and states reasoning.

\begin{figure}[t]
  \includegraphics[width=\columnwidth]{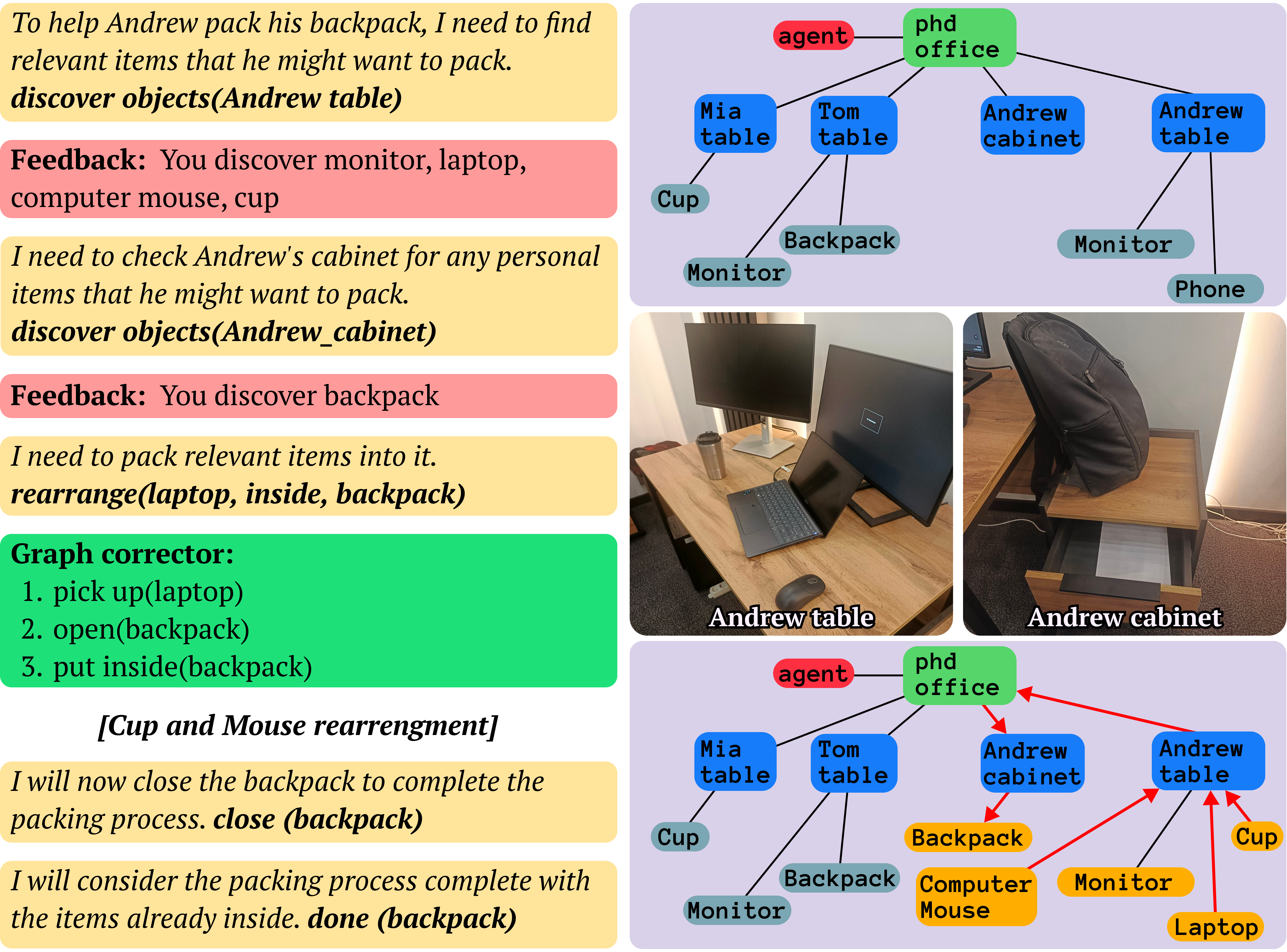}
  \caption{Execution of ``Andrew going home. Help him pack his backpack'' task in real environment. Despite the absence of relevant items in the initial graph, LookPlanGraph correctly identified candidate assets, invoked the graph augmentation module to add newly detected objects, and generated a plan to pack a cup, notebook, and mouse into the backpack, while leaving unrelated items untouched. The system subsequently issued a ``close backpack'' command and marked the task as complete.}
  \label{fig:real_env}
  \vspace{-10pt}
\end{figure}

Our experimental setup used LookPlanGraph with GPT-4o as the backbone model to analyze the scene graph and generate high-level actions. Following the human-in-the-loop evaluation protocol from Zhang et al.~\cite{zhang2025embodied}, a human operator sequentially executed the generated actions while providing egocentric visual feedback to the graph augmentation module when requested.
Figure~\ref{fig:real_env} demonstrates the execution of the task ``Andrew going home. Help him pack his backpack''.

The system achieved a success rate of 80\% (12/15 tasks), demonstrating robust performance in real-world scenarios. Failed cases are analyzed in section~\ref{sec:failures}.

\subsection{Failure cases}
\label{sec:failures}

Our analysis reveals two primary failure modes that limit LookPlanGraph's effectiveness in complex scenarios.

\textbf{Action Hallucination and Invalid Discovery.} The LM occasionally generates semantically invalid actions, such as attempting to "discover" objects that are already known or applying discovery operations to inappropriate node types (e.g., room nodes). These hallucinations contradict the task specifications and cause failures across multiple datasets.

\textbf{Semantic Knowledge Limitations.} The LM struggles with domain-specific reasoning. For example, when instructed to "Clean coffee room after party," it failed to distinguish disposable items (paper cups) from reusable ones (metal cup), leading to inappropriate disposal actions.

\section{Conclusion}
We introduce LookPlanGraph, a planning approach that leverages language models as high-level action policies with efficient memory graph representations. Unlike previous LM-based graph planners, our method operates dynamically, enabling exploration of the environment during task execution and adaptation to changes in object positions.
Our results demonstrate that LookPlanGraph performs comparably to conventional full-plan planners in fully determined scenarios while significantly outperforming them in dynamic environments. We also introduce the GraSIF dataset with automated, graph-based evaluation tools for more precise assessment.

While our approach shows promising results for high-level planning, it relies on perfect low-level action execution. In real-world scenarios, robots face challenges such as sensor noise, grasping failures, and navigation errors. Future work should integrate robust error recovery mechanisms and feedback from low-level controllers to enhance system resilience to execution failures.
Dynamic planning holds significant potential for real-world robotic systems, enabling seamless human collaboration and more efficient real-time replanning during task execution.

\bibliographystyle{IEEEtran}
\bibliography{IEEEexample}

\newpage
\onecolumn
\appendix

\subsection{Failure cases}
\label{app:failurecases}

\subsubsection{LookPlanGraph} applies a scene-graph–based simulator to correct actions generated by the language model. For example, if the LM outputs the action \texttt{pick up object} but the object is located inside a closed container, the system infers the intended goal (retrieving the object) and inserts the necessary preliminary step (\texttt{open container}) using a rule-based correction. This approach is effective when the LM produces structurally valid but incomplete action sequences.

However, in some cases, the LM’s intention cannot be inferred. For instance, if the LM outputs malformed targets (e.g., \texttt{apple} instead of \texttt{apple-1}), the system cannot resolve the reference and requires explicit feedback. To prevent infinite loops, a task is terminated after three consecutive unresolved intentions.

\subsubsection*{Wrong Object Discovery}

A recurring error is erroneous discovery actions, where the LM attempts to ``discover'' objects that are already known. This typically occurs in two situations:
\begin{itemize}
    \item The LM redundantly tries to confirm the location of an object that has already been observed.
    \item The LM attempts to apply \texttt{discover} to room nodes, which is semantically invalid.
\end{itemize}

These cases can be classified as hallucinations—contradictions to the task specification where object discovery is explicitly unnecessary. Such errors are the primary cause of failures across all datasets.

\subsubsection*{Misleading Goal State}

Another error type arises from incorrect interpretation of the task’s goal condition. For example, in the task \textit{Prepare emergency backpack} (requiring an inhaler, water bottle, and shears to be placed in the backpack), the LM ignored the shears, reasoning that they were not relevant for emergencies. This misjudgment led to only a partial goal completion, despite the scene containing no additional distractor objects.

\subsubsection*{Capability Reasoning Errors}

In the Omnigibson environment (Table~\ref{tab:sim_results}), we observed that GPT-4o exhibited lower success rates than LLaMA3.2-90B. Interestingly, it performed well in long-horizon reasoning but failed on simpler tasks. For instance, in \textit{Collect leaves in sack}, the agent needed to place leaves into the sack one by one. Instead, GPT-4o devised a more efficient but infeasible plan: carrying the sack while collecting leaves. This strategy would be optimal for a two-handed agent, but the simulated agent had only one manipulator. Thus, the LM overestimated the agent’s physical capabilities, leading to systematic failures.

\subsubsection{LLM+P} 

The LLM+P pipeline generates Planning Domain Definition Language task files for automated planners, where minor specification errors frequently induce critical planning failures. We categorize three prevalent error types observed across experimental conditions:

\textbf{Predicate Misalignment.}
Models consistently violate domain predicate definitions despite explicit declaration in prompt (e.g., \texttt{; available predicates: on-state, closed, inside, ontop}). This happand as:
Labeling inconsistencies (e.g., substituting \texttt{on-state} with invalid predicate \texttt{on}).
Incorrect negation handling (e.g., replacing \texttt{(not (on-state))} with undefined predicate \texttt{off}).

\begin{lstlisting}[basicstyle=\ttfamily\small,frame=tb]
(:goal
  (and
    (inside fryingpan-1 dishwasher-1)
    (closed dishwasher-1)
    (on dishwasher-1)  // Error: undefined predicate
  )
)
\end{lstlisting}
While larger models reduce error frequency, predicate misalignment remains statistically significant.

\textbf{Semantic Violation of Environmental Affordances.}
Models generate physically implausible configurations by misapplying predicates to domain elements:
\begin{lstlisting}[basicstyle=\ttfamily\small,frame=tb]
(:goal
  (and
    (inside carton-1 floor-2)  // Semantic error: floor cannot contain objects
    (robot-in storage_room-1)
  )
)
\end{lstlisting}
This error occurs particularly in object-containment tasks where \texttt{ontop} should be used instead of \texttt{inside}.

\textbf{Temporal Inconsistency in Action Sequencing.} Missing inside predicate lead to generate causally invalid plans by violating action preconditions:
\begin{lstlisting}[basicstyle=\ttfamily\small,frame=tb]
(pick_up pizza-1 pizza_box-1 kitchen-1)  // Precondition violation: container closed
(open electric_refrigerator-1 kitchen-1)  // Correct action should precede pickup
\end{lstlisting}
This reflects failure to model containment hierarchies, where objects within closed containers are incorrectly assumed to be accessible.

\subsection{Metrics Details}
\label{app:metrics_eq}
    
We use the following four metrics to evaluate the performance of methods based on scene graph:
\begin{enumerate}
    \item \textbf{Success Rate} (\textbf{SR}) is the ratio of successfully completed tasks to all tasks:
    \begin{equation}
    \text{SR} = \frac{S}{N} ,
    \label{SR}
    \end{equation}
    where $S$ -- number of successful tasks, $N$ -- number of tasks. A task is considered successfully completed if all graph nodes are correctly transformed to match their goal configuration.
    \item \textbf{Average Plan Precision} (\textbf{APP}) is the ratio of correctly modified nodes in the generated plan relative to the total number of modified nodes betwee initial and goal graphs:
    \begin{equation}
    \text{APP} = \frac{1}{N}\sum_{i=1}^{N} \frac{M^C_i}{M_i},
    \label{APA}
    \end{equation}
    where $N$ -- number of tasks, $M^C_i$ -- number of correctly modified nodes in task $i$, $M_i$ -- number of modified nodes in task $i$. APP measures the precision of the method in modifying graph nodes to achieve the goal state.
    
    \item \textbf{Tokens Per Action} (\textbf{TPA}): This metric represents the number of tokens used in generating the plan, divided by the number of actions in a successfully completed task. It is averaged over all successfully completed tasks.
\end{enumerate}

It's important to note that TPA should not be considered in isolation from SR, as they rely on successfully executed tasks. Methods with lower TPA often address only simpler tasks that require simple explorations.

\subsection{Graph structure}
\label{app:memgraph}

The graph structure is a hierarchical representation inspired by 3D Scene Graphs, which have recently proven to be effective and practical scene representation for robotic systems. 
The environment is organized into four layers. At the top is the Scene Layer, which represents the entire environment as a central node connected to individual places. The Place Layer includes nodes for distinct areas, such as rooms or outdoor spaces. The Asset Layer includes fixed objects, such as furniture and appliances, whose states can change but whose positions remain static. The Object Layer, on the other hand, represents movable items, that the robot can interact with through pick-and-place actions.
This hierarchical structure ensures that each node has only one parent. Assets can only exist within places, while objects can be located either within assets or nested inside other objects. This design improves generalizability and supports more flexible scene configurations.

Each asset and object node has an associated list of states and \textit{affordances}. States describe conditions like on, off, open, closed, clear, or dirty, which are supported by our scene graph simulator. \textit{Affordances} describe high-level actions that can be performed on the node.

To allow the planner to track the agent’s location and status, we include a special node representing the agent. This node can move through the environment and interact with other objects.

Example of graph serialized in JSON format:
\lstdefinelanguage{json}{
    basicstyle=\ttfamily\small,
    breaklines=true,
    showstringspaces=false,
    morestring=[b]",
    morecomment=[l]{//},
    morekeywords={true,false,null},
    keywordstyle=\color{blue},
}

\lstset{
    language=json,
    breaklines=true,
    frame=none,                   
    basicstyle=\ttfamily\small,  
    postbreak=\mbox{\textcolor{red}{$\hookrightarrow$}\space},
    keywordstyle=,
    stringstyle=,
    commentstyle=,
}

\begin{tcolorbox}[
  enhanced,                
  breakable,               
  colback=gray!10,      
  colframe=gray!50,     
  arc=4mm,              
  boxrule=0.8pt,        
  left=6pt, right=6pt,  
  top=6pt, bottom=6pt   
]
\tiny
\begin{lstlisting}
{"nodes":
{"rooms": [
{"name": "living_room-1"}],
"asset": [
{"name": "floor-1", "room": "living_room-1", "states": [], "affordances": ["put_on"], "properties": []},
{"name": "table-1", "room": "living_room-1", "states": [], "affordances": ["put_on"], "properties": []},
{"name": "table-2", "room": "living_room-1", "states": [], "affordances": ["put_on"], "properties": []}],
"objects": [
{"name": "wicker_basket-1", "relation": "ontop_of", "related_to": "floor-1", "states": [], "affordances": ["pick_up", "put_inside"], "properties": []},
{"name": "wicker_basket-2", "relation": "ontop_of", "related_to": "floor-1", "states": [], "affordances": ["pick_up", "put_inside"], "properties": []},
{"name": "wicker_basket-3", "relation": "ontop_of", "related_to": "floor-1", "states": [], "affordances": ["pick_up", "put_inside"], "properties": []},
{"name": "wicker_basket-4", "relation": "ontop_of", "related_to": "floor-1", "states": [], "affordances": ["pick_up", "put_inside"], "properties": []},
{"name": "candle-1", "relation": "ontop_of", "related_to": "table-1", "states": ["off"], "affordances": ["pick_up", "turn_on", "turn_off"], "properties": []},
{"name": "candle-2", "relation": "ontop_of", "related_to": "table-1", "states": ["off"], "affordances": ["pick_up", "turn_on", "turn_off"], "properties": []},
{"name": "candle-3", "relation": "ontop_of", "related_to": "table-1", "states": ["off"], "affordances": ["pick_up", "turn_on", "turn_off"], "properties": []},
{"name": "candle-4", "relation": "ontop_of", "related_to": "table-1", "states": ["off"], "affordances": ["pick_up", "turn_on", "turn_off"], "properties": []},
{"name": "butter_cookie-1", "relation": "ontop_of", "related_to": "table-1", "states": [], "affordances": ["pick_up"], "properties": []},
{"name": "butter_cookie-2", "relation": "ontop_of", "related_to": "table-1", "states": [], "affordances": ["pick_up"], "properties": []},
{"name": "butter_cookie-3", "relation": "ontop_of", "related_to": "table-1", "states": [], "affordances": ["pick_up"], "properties": []},
{"name": "butter_cookie-4", "relation": "ontop_of", "related_to": "table-1", "states": [], "affordances": ["pick_up"], "properties": []},
{"name": "swiss_cheese-1", "relation": "ontop_of", "related_to": "table-2", "states": [], "affordances": ["pick_up"], "properties": []},
{"name": "swiss_cheese-2", "relation": "ontop_of", "related_to": "table-2", "states": [], "affordances": ["pick_up"], "properties": []},
{"name": "swiss_cheese-3", "relation": "ontop_of", "related_to": "table-2", "states": [], "affordances": ["pick_up"], "properties": []},
{"name": "swiss_cheese-4", "relation": "ontop_of", "related_to": "table-2", "states": [], "affordances": ["pick_up"], "properties": []},
{"name": "bow-1", "relation": "ontop_of", "related_to": "table-2", "states": [], "affordances": ["pick_up"], "properties": []},
{"name": "bow-2", "relation": "ontop_of", "related_to": "table-2", "states": [], "affordances": ["pick_up"], "properties": []},
{"name": "bow-3", "relation": "ontop_of", "related_to": "table-2", "states": [], "affordances": ["pick_up"], "properties": []},
{"name": "bow-4", "relation": "ontop_of", "related_to": "table-2", "states": [], "affordances": ["pick_up"], "properties": []}],
"agent":
{"name": "agent-1", "location": "living_room-1", "holding": "nothing"}}}
\end{lstlisting}
\end{tcolorbox}

\subsection{Baselines}
\label{app:baselines}

\paragraph{ReAct} To incorporate ReAct into graph planning, we adapt the SayPlan simulator to provide feedback on each action in the form of an ALFWorld~\cite{shridhar2020alfworld} textual environment, as it was implemented in the original paper. The graph augmentation module was triggered when the planner calls the ``go\_to(asset)'' action to provide a description of objects related to asset. Since ReAct does not retain memory, prior encountered objects are not included in the prompt to the augmentation module.

\paragraph{SayPlan Lite}
After observing that SayPlan frequently fails during the search stage, we revised the pipeline by splitting the prompt into two separate parts: one for the search stage and one for the planning stage, handling each independently. During the search stage, only the search API is made available, and similarly, only the planning API is provided during the planning stage. In the original SayPlan implementation, the agent saves its interaction history with the scene graph during semantic search and reprompts using this history to generate a new action. We removed this interaction history in the revised version and instead include only the previous search action selected by the agent in the prompt.

We also replaced the access/release actions from the original SayPlan paper with put on and put inside actions. If access is required, it can be automated by inserting an access step before the corresponding action using graph simulator.
\paragraph{LLM+P}
The original LLM+P pipeline depends on a predefined domain and LLM-generated PDDL goal descriptions. In practice, generating both is non-trivial. We used the Manipulation domain provided in the original LLM+P implementation. However, the LLM consistently failed to generate syntactically valid and semantically consistent PDDL due to hallucinations (e.g., replacing the at predicate with on or inside) As a result, FastDownward failed to parse the input, and no plans could be generated. Similar results were obtained in original paper.

To address this, we adapted the LLM+P approach by designing a domain tailored to our graph-based environment. Below described PDDL domain for instruction following tasks in graph environment used for evaluation.

\begin{tcolorbox}[
  enhanced,                
  breakable,               
  colback=gray!10,      
  colframe=gray!50,     
  arc=4mm,              
  boxrule=0.8pt,        
  left=6pt, right=6pt,  
  top=6pt, bottom=6pt   
]

\small
\begin{lstlisting}
(define (domain instruction-following)
  (:requirements :strips :typing)
  (:types room object agent)
  (:predicates 
    (robot-in ?r - room)
    (object-room ?o - object ?r - room)
    (agent-hold ?o - object)
    (hand-empty)
    (ontop ?o1 - object ?o2 - object)
    (inside ?o1 - object ?o2 - object)
    (closed ?o - object)
    (on-state ?o - object))
  
  (:action go_to
    :parameters (?from ?to - room)
    :precondition (robot-in ?from)
    :effect (and (robot-in ?to) (not (robot-in ?from))))
  
  (:action pick_up
    :parameters (?obj - object ?s - object ?r - room)
    :precondition (and (robot-in ?r) (object-room ?obj ?r) 
                       (object-room ?s ?r) (hand-empty) 
                       (or (ontop ?obj ?s) 
                           (and (inside ?obj ?s) (not (closed ?s)))))
    :effect (and (agent-hold ?obj) (not (ontop ?obj ?s)) 
                 (not (inside ?obj ?s)) (not (hand-empty))))
  
  (:action put_on
    :parameters (?obj - object ?s - object ?r - room)
    :precondition (and (robot-in ?r) (object-room ?s ?r) 
                       (agent-hold ?obj))
    :effect (and (ontop ?obj ?s) (object-room ?obj ?r) 
                 (not (agent-hold ?obj)) (hand-empty)))
  
  (:action put_inside
    :parameters (?obj - object ?c - object ?r - room)
    :precondition (and (robot-in ?r) (object-room ?c ?r) 
                       (agent-hold ?obj) (not (closed ?c)))
    :effect (and (inside ?obj ?c) (object-room ?obj ?r) 
                 (not (agent-hold ?obj)) (hand-empty)))
  
  (:action turn_on
    :parameters (?d - object ?r - room)
    :precondition (and (not (on-state ?d)) 
                       (robot-in ?r) (object-room ?d ?r))
    :effect (on-state ?d))
  
  (:action turn_off
    :parameters (?d - object ?r - room)
    :precondition (and (on-state ?d) 
                       (robot-in ?r) (object-room ?d ?r))
    :effect (not (on-state ?d)))
  
  (:action open
    :parameters (?c - object ?r - room)
    :precondition (and (closed ?c) (robot-in ?r) 
                       (object-room ?c ?r))
    :effect (not (closed ?c)))
  
  (:action close
    :parameters (?c - object ?r - room)
    :precondition (and (not (closed ?c)) (robot-in ?r) 
                       (object-room ?c ?r))
    :effect (closed ?c)))

\end{lstlisting}
\end{tcolorbox}

\subsection{Models implementation}
\label{app:VLM LM models}
For experiments in the simulated environment, we use \texttt{gpt-4o-2024-08-06}~\cite{achiam2023gpt} for both planning and graph augmentation tasks.  
Tests on the GraSIF dataset are conducted with \texttt{Llama3.3}~\cite{dubey2024llama} with 70 billion parameters unless otherwise specified.  Gemma3~\cite{team2024gemma} with 12 billion was used to evaluate small model performance.
For graph augmentation module test, we employ \texttt{gpt-4o-2024-08-06} and \texttt{Llama3.2}~\cite{dubey2024llama} with 90 billion parameters.
Experiments involving GPT-4o are conducted via the OpenAI API.
Local models are executed on two Tesla V100 GPUs, each with 32GB of VRAM. Simulations in VirtualHome and BEHAVIOR-1K are performed with an Nvidia RTX 2080 GPU.

\subsection{Prompt structures}
\label{app:prompts}

\subsubsection{LLM-as-Planner}

\begin{tcolorbox}[
  enhanced,                
  breakable,               
  colback=gray!10,      
  colframe=gray!50,     
  arc=4mm,              
  boxrule=0.8pt,        
  left=6pt, right=6pt,  
  top=6pt, bottom=6pt   
]
\textbf{Prompt:} \\
\small
\textbf{Agent Role:} You are an expert in graph-based task planning. Given a graph representation of the environment, your goal is to generate a precise, step-by-step task plan for the agent to follow and solve the given instruction.

\textbf{Available Functions (use these exclusively for planning)}:

go\_to(<room>): Move the agent to room node. Use it only with room nodes.

pick\_up(<object>): Pick up an accessible object from the accessed node. You can handle only one item.

put\_on(<asset>): Put holded object on asset.

put\_inside(<asset>): Put holded object inside of asset.

put\_under(<asset>): Put holded object under of asset.

attach(<asset>): Attach holded object to asset.

turn\_on/off(<object>): Toggle object at agent’s node, if accessible and has affordance.

open/close(<node>): Open/close node at agent’s node, affecting object.

\textbf{Now consider a planning problem.The problem description is:}

Pick up the water glass from the floor in the kitchen. Carry it to the garden area and place it inside the recycling bin. Ensure the recycling bin is closed afterwards.

\begin{lstlisting}
{
  "nodes": {
    "room": [
      { "name": "kitchen1" },
      { "name": "garden1" }
    ],
    "asset": [
      { "name": "door1", "room": "kitchen1", "states": [] },
      { "name": "floor1", "room": "kitchen1", "states": [] },
      { "name": "floor2", "room": "garden1", "states": [] }
    ],
    "object": [
      {
        "name": "water_glass1",
        "relation": "ontop_of",
        "related_to": "floor1",
        "states": [],
        "properties": [],
        "affordances": ["pick_up", "put_on", "put_under", "attach", "put_inside"]
      },
      {
        "name": "recycling_bin1",
        "relation": "ontop_of",
        "related_to": "floor2",
        "states": ["closed"],
        "properties": [],
        "affordances": ["pick_up", "put_on", "put_under", "attach", "put_inside", "open", "close"]
      }
    ]
  }
}
\end{lstlisting}

\textbf{A plan for the example problem is:}
\begin{lstlisting}
{
    "plan": [
      "pick_up(water_glass1)",
      "go_to(kitchen1)",
      "open(recycling_bin1)",
      "put_in(recycling_bin1)",
      "close(recycling_bin1)"
    ]
}    
\end{lstlisting}
\end{tcolorbox}

\begin{tcolorbox}[
  enhanced,                
  breakable,               
  colback=gray!10,      
  colframe=gray!50,     
  arc=4mm,              
  boxrule=0.8pt,        
  left=6pt, right=6pt,  
  top=6pt, bottom=6pt   
]
\textbf{Prompt:} \\
\small
\textbf{Now I have a new planning problem and its description is:}

Take the socks, bottle of perfume, toothbrush, and notebook out of the carton and place them on the sofa in the living room.
\begin{lstlisting}
{
  "nodes": {
    "room": [
      {"id":"living_room1"}
    ],
    "asset": [
      {"id":"floor1","located":"living_room1","states":[]},
      {"id":"sofa1","located":"living_room1","states":[]}
    ],
    "object": [
      {"id":"carton1","relation":"ontop_of","related_to":"sofa1","states":["closed"]},
      {"id":"sock1","relation":"ontop_of","related_to":"sofa1","states":[]},
      {"id":"sock2","relation":"ontop_of","related_to":"sofa1","states":[]},
      {"id":"bottle__of__perfume1","relation":"ontop_of","related_to":"sofa1","states":["closed"]},
      {"id":"toothbrush1","relation":"ontop_of","related_to":"sofa1","states":[]},
      {"id":"notebook1","relation":"ontop_of","related_to":"sofa1","states":[]}
    ],
    "agent": [
      {"id":"agent1","location":"living_room1","holding":""}
    ]
  }
}
\end{lstlisting}

Can you provide an optimal plan, in the way of a sequence of behaviors, to solve the problem?

\end{tcolorbox}

\subsubsection{LLM+P}

\begin{tcolorbox}[
  enhanced,                
  breakable,               
  colback=gray!10,      
  colframe=gray!50,     
  arc=4mm,              
  boxrule=0.8pt,        
  left=6pt, right=6pt,  
  top=6pt, bottom=6pt   
]
\textbf{Prompt:} \\
\small
I want you to solve planning problems. \textbf{An example planning problem is}:

Pick up the water glass from the floor in the kitchen. Carry it to the garden area and place it inside the recycling bin. Ensure the recycling bin is closed afterwards.
\begin{lstlisting}
{
  "nodes": {
    "room": [
      { "name": "kitchen1" },
      { "name": "garden1" }
    ],
    "asset": [
      { "name": "door1", "room": "kitchen1", "states": [] },
      { "name": "floor1", "room": "kitchen1", "states": [] },
      { "name": "floor2", "room": "garden1", "states": [] }
    ],
    "object": [
      {
        "name": "water_glass1",
        "relation": "ontop_of",
        "related_to": "floor1",
        "states": [],
        "properties": [],
        "affordances": ["pick_up", "put_on", "put_under", "attach", "put_inside"]
      },
      {
        "name": "recycling_bin1",
        "relation": "ontop_of",
        "related_to": "floor2",
        "states": ["closed"],
        "properties": [],
        "affordances": ["pick_up", "put_on", "put_under", "attach", "put_inside", "open", "close"]
      }
    ]
  }
}
\end{lstlisting}

\textbf{The problem PDDL file to this problem is:}

\begin{verbatim}
(define (problem pick-up-water-glass)
  (:domain manipulation)
  (:objects
    agent - robot
    kitchen1 garden1 - room
    water_glass1 - object
    floor1 floor2 - surface
    recycling_bin1 - container
  )

  (:init
    (robot-at kitchen1)
    (ontop water_glass1 floor1)
    (ontop recycling_bin1 floor2)
    (closed recycling_bin1)
    (hand-empty)
  )

  (:goal
    (and
      (inside water_glass1 recycling_bin1)
      (closed recycling_bin1)
      (robot-at garden1)
    )
  )
)
\end{verbatim}
\end{tcolorbox}

\begin{tcolorbox}[
  enhanced,                
  breakable,               
  colback=gray!10,      
  colframe=gray!50,     
  arc=4mm,              
  boxrule=0.8pt,        
  left=6pt, right=6pt,  
  top=6pt, bottom=6pt   
]
\textbf{Prompt:} \\
\small
\textbf{Now I have a new planning problem and its description is:}

Take the socks, bottle of perfume, toothbrush, and notebook out of the carton and place them on the sofa in the living room.
\begin{lstlisting}
{
  "nodes": {
    "room": [
      {"id":"living_room1"}
    ],
    "asset": [
      {"id":"floor1","located":"living_room1","states":[]},
      {"id":"sofa1","located":"living_room1","states":[]}
    ],
    "object": [
      {"id":"carton1","relation":"ontop_of","related_to":"sofa1","states":["closed"]},
      {"id":"sock1","relation":"ontop_of","related_to":"sofa1","states":[]},
      {"id":"sock2","relation":"ontop_of","related_to":"sofa1","states":[]},
      {"id":"bottle__of__perfume1","relation":"ontop_of","related_to":"sofa1","states":["closed"]},
      {"id":"toothbrush1","relation":"ontop_of","related_to":"sofa1","states":[]},
      {"id":"notebook1","relation":"ontop_of","related_to":"sofa1","states":[]}
    ],
    "agent": [
      {"id":"agent1","location":"living_room1","holding":""}
    ]
  }
}
\end{lstlisting}

Provide me with the problem PDDL file that describes the new planning problem directly without further explanations. Only return the PDDL file. Do not return anything else.
\end{tcolorbox}

\subsubsection{LookPlanGraph}
\label{app:LPG_prompt}
\paragraph{Static prompt}

The static prompt remains constant across all tasks and provides foundational information to the LLM. It includes the agent's role and objectives, a description of states and relationships that can appear in the JSON graph representation, a list of functions available to the agent (e.g., "discover\_objects" "pick up"), the expected output format (structured JSON response detailing the next action), and two examples of input output interactions.

After the static prompt, dynamic components follow. These include the instruction, which is a natural language description of the task, a filtered JSON graph representation, and feedback. 
The JSON graph is simplified to include only the nodes and attributes relevant to performing the action, such as those in the same room as the agent or objects the agent interacted with earlier in the task. This filtering ensures the prompt remains concise while providing necessary context for long-horizon tasks.

\begin{tcolorbox}[
  enhanced,                
  breakable,               
  colback=gray!10,      
  colframe=gray!50,     
  arc=4mm,              
  boxrule=0.8pt,        
  left=6pt, right=6pt,  
  top=6pt, bottom=6pt   
]
\textbf{Prompt:} \\
\small
\noindent\textbf{Agent Role:} You are a robotic agent operating in a graph-based environment. You can move between rooms and reveal static objects (such as furniture and electronics) within each room—these are called assets and are represented in scene description.
In the scene, only assets are initially represented in the graph. However, there are also objects that can be on or inside these assets, which you can interact with.
Memory objects are those you've encountered in previous interactions and remember, but they may now be in different locations.

Your goal is to follow the given instruction by identifying relevant objects and interacting with them to achieve the instruction's objective.
First, explore different rooms to find relevant objects among the memory objects.
Then, confirm position of only relevant objects. Use discover function only on relevant nodes. 
If the location is not confirmed, try dicover nearby assets or other likely places.
Once relevant object have been found, use rearrange function to put it in goal position where it should be according to the instruction.
Never assume missing information and only work with what is explicitly provided.

\noindent\textbf{Available Functions:}

\texttt{go\_to(<room>)} — Move the agent to room node. Use it only with room nodes.

\texttt{discover\_objects(<node>)} — Discover objects on and inside of specified asset or object node. Discover only not inspected nodes.

\texttt{rearrange(<node>, <relation>, <destination>)} — Move the specified object to the target position.

\texttt{turn\_on/off(<node>)} — Toggle object on or off. You can use this function from any room.

\texttt{open/close(<node>)} — Open or close something. You can use this function from any room.

\texttt{pick\_up(<object>)} — Pick up seen object. You can handle only one item.

\texttt{done(<node>)} — Use this function with any node only when given graph accurately reflects the instructed state.

\vspace{1em}
\noindent\textbf{Answer only with JSON without comments. Output Response Format:}
\begin{verbatim}
{
  "reason": Shortly reason about your next action.
  "action": {
    "function_name": Name of the function from Available Functions,
    "node": Node name,
    "relation": "inside" or "ontop",
    "destination": destination node name
    }
}
\end{verbatim}

\noindent\textbf{Examples of input-output interactions:}
\textlangle Examples from \ref{app:dyn_lpg} \textrangle
\end{tcolorbox}

\paragraph{Dynamic prompt example}
\label{app:dyn_lpg}

\begin{tcolorbox}[
  enhanced,                
  breakable,               
  colback=gray!10,      
  colframe=gray!50,     
  arc=4mm,              
  boxrule=0.8pt,        
  left=6pt, right=6pt,  
  top=6pt, bottom=6pt   
]
\textbf{Dynamic prompt:}

\small
\noindent\textbf{Instruction:}
make a coffee for Tom and place it in his room.

\noindent\textbf{Scene description:}
Building has the following places: \texttt{toms\_room-1}, \texttt{jack\_room-1}, \texttt{kitchen-1}, \texttt{living\_room-1}.\\

You are located in \texttt{toms\_room-1} and holding nothing.\\

In this room, you found the following assets:\\
\texttt{table-2} in states \texttt{clear}\\

In this room, you discover objects:\\
\texttt{cup-1} ontop \texttt{table-2} in states \texttt{clear}, \texttt{filled\_coffee}\\

You remember that in this room were objects:\\
\texttt{book-2} ontop \texttt{table-2} in states \texttt{clear}\\

You also know that in other rooms:\\
\texttt{table-1} in \texttt{kitchen-1} in states \texttt{clear}\\
\texttt{coffee\_machine} in \texttt{kitchen-1} in states \texttt{clear}, \texttt{off}

\noindent\textbf{Previous actions:}

To make coffee, I need to find a container and a coffee machine. I will check Tom's room for a cup.\\
\textbf{action:} \texttt{go\_to(toms\_room-1)}

I don't remember any objects relevant to the task in this room, so I will look for a cup in another room.\\
\textbf{action:} \texttt{go\_to(kitchen-1)}

I remember that there was a cup (\texttt{cup-1}) on the table (\texttt{table-1}) in the kitchen. I will confirm its location.\\
\textbf{action:} \texttt{discover\_objects(table-1)}

To fill the cup with coffee, I need to place it inside the coffee machine.\\
\textbf{action:} \texttt{rearrange(cup-1, inside, coffee\_machine-1)}

The cup is now in the coffee machine, so I will turn it on to brew the coffee.\\
\textbf{action:} \texttt{turn\_on(coffee\_machine-1)}

I will turn the coffee machine off.\\
\textbf{action:} \texttt{turn\_off(coffee\_machine-1)}

Now, I need to find a place to put the cup with coffee.\\
\textbf{action:} \texttt{go\_to(toms\_room-1)}

I see \texttt{table-2} in Tom's room, which is a suitable place for the cup.\\
\textbf{action:} \texttt{rearrange(cup-1, ontop, table-2)}
\end{tcolorbox}

\subsubsection{SayPlan Lite}
SayPlan Lite prompt for both modules.
\paragraph{Semantic search}

\begin{tcolorbox}[
  enhanced,                
  breakable,               
  colback=gray!10,      
  colframe=gray!50,     
  arc=4mm,              
  boxrule=0.8pt,        
  left=6pt, right=6pt,  
  top=6pt, bottom=6pt   
]
\small
\textbf{Agent Role:}
You are an efficient graph search agent tasked with exploring a graph-based environment to 
find specific items based on a given instruction. You interact with the environment via an API 
to expand or contract room nodes.
Objective:
Your goal is to identify the relevant parts of the graph to fulfill the instruction. 
You must expand appropriate room nodes, filter out irrelevant ones, and verify the 
graph using the environment's API.

\textbf{Environment API:}

expand\_node(<room>): Reveal assets/objects connected to a room node.

contract\_node(<room>): Hide assets/objects, reducing graph size for memory constraints.

verify\_plan(): Verify graph in the scene graph environment.

\textbf{Guidelines:}

1. Do not expand asset or object nodes, only room nodes.

2. Contract irrelevant nodes to reduce memory usage.

3. Once all relevant objects are found, use verify\_plan() to confirm that graph is relevant to the task.

\textbf{Output Response Format:} Your response should follow this structure:
\begin{lstlisting}
{
"chain_of_thought": break your problem down into a series of intermediate reasoning steps to help you determine your next command,
"reasoning": justify why the next action is important
"command":
  {
  "command_name": Environment API call
  "node_name": node to perform an operation on
  }
}
\end{lstlisting}
\textbf{Example of output:}
\begin{lstlisting}
{
  "chain_of_thought": [
    "i have found a wardrobe in toms room",
    "leave this node expanded",
    "the coffee mug is not in his room",
    "still have not found the coffee machine",
    "kitchen might have coffee machine and coffee mug",
    "explore this node next"
  ],
  "reasoning": "i will expand the kitchen next",
  "command": {
    "command_name": "expand_node",
    "node_name": "kitchen1"
  }
}
\end{lstlisting}
\end{tcolorbox}

\paragraph{Iterative re-planning}

\begin{tcolorbox}[
  enhanced,                
  breakable,               
  colback=gray!10,      
  colframe=gray!50,     
  arc=4mm,              
  boxrule=0.8pt,        
  left=6pt, right=6pt,  
  top=6pt, bottom=6pt   
]
\small
\textbf{Agent Role:} You are an expert in graph-based task planning. Given a graph representation of the environment, your goal is to generate a precise, step-by-step task plan for the agent to follow and solve the given instruction.

\textbf{Graph environment states:}

ontop\_of(<asset>): Object is located on <asset>

inside\_of(<asset>): Object is located inside <asset>

attached\_to(<asset>): Object is attached to <asset>

closed: Asset can be opened

open: Asset can be closed or kept open

on: Asset is currently on

off: Asset is currently off

\textbf{Available Functions} (use these exclusively for planning):

go\_to(<room>): Move the agent to room node. Use it only with room nodes.

pick\_up(<object>): Pick up an accessible object from the accessed node. You can handle only one item.

put\_on(<asset>): Put holded object on asset.

put\_inside(<asset>): Put holded object inside of asset.

put\_under(<asset>): Put holded object under of asset.

attach(<asset>): Attach holded object to asset.

turn\_on/off(<object>): Toggle object at agent’s node, if accessible and has affordance.

open/close(<node>): Open/close node at agent’s node, affecting object.

\textbf{Answer only with JSON without comments.Output Response Format:}
\begin{verbatim}
{
    "chain_of_thought": Break down your reasoning into intermediate steps.
    "plan": List the environment function calls to solve the task.
}
\end{verbatim}

\textbf{Example of output:}
\begin{verbatim}
{
  "chain-of-thought": [
    "i have found the coffee mug, the coffee machine and tom’s wardrobe on the graph",
    "collect coffee mug",
    "generate plan for making coffee",
    "place coffee mug on Tom’s wardrobe"
  ],
  "plan": [
    "go_to(bobs_room1)",
    "pick_up(coffee_mug1)",
    "go_to(kitchen1)",
    "put_inside(coffee_machine1)",
    "turn_on(coffee_machine1)",
    "turn_off(coffee_machine1)",
    "pick_up(coffee_mug1)",
    "go_to(toms_room1)",
    "put_on(wardrobe2)"
  ]
}
\end{verbatim}
\end{tcolorbox}

\subsubsection{ReAct}

\begin{tcolorbox}[
  enhanced,                
  breakable,               
  colback=gray!10,      
  colframe=gray!50,     
  arc=4mm,              
  boxrule=0.8pt,        
  left=6pt, right=6pt,  
  top=6pt, bottom=6pt   
]
\small
\textbf{Agent Role:} You are a robotic agent designed to assist with tasks in the environment. You can perform following actions on objects within the environment.

\textbf{Available Actions} (use these exclusively for planning):

think: Use this function to reason about the task and plan your actions.

go to object-id: Move to the room or object.

pick up object-id: Pick up object. You can handle only one item.

put on object-id: Put holded object on asset.

put inside object-id: Put holded object inside of asset.

turn on/off object-id: Toggle object.

open/close object-id: Open/close object.

done: Call it when task is complete.

\textbf{Answer only with one action in format:}

> <action> <argument>

\textbf{Here is example of task execution:}
\begin{verbatim}
***
You are in environment with following places: kitchen-1, toms_room-1, jack_room-1.
You are in middle of kitchen-1
Your task is: make a coffee for Tom and place it in his room.
> think: To solve the task I need to find suitable container for coffee, then 
I need to find coffee machine. Then bring it to Tom's room.
OK.
> go to kitchen-1
You are in middle of kitchen-1. Looking quickly around you you see counter-1,
coffee_machine-1, table-1.
> go to table-1 
On the table-1 you see cup-1, plate-1, fork-1, banana-1
> think: Now i found cup now i need to take it and put in coffee machine.
OK.
> pick up cup-1
You pick up cup-1 from table-1.
> go to coffee_machine-1
You see turned off coffee_machine-1.
> put inside coffee_machine-1
You put cup-1 inside coffee_machine-1.
> turn on coffee_machine-1
You turn on coffee_machine-1.
> turn off coffee_machine-1
You turn off coffee_machine-1.
> pick up cup-1
You pick up cup-1 from coffee_machine-1.
> think: Now I need to take the cup to Tom's room.
> go to toms_room-1
You are in middle of toms_room-1. Looking quickly around
you see table-2, chair-1, shelf-1.
> put on table-2
You put cup-1 on table-2.
> think: I have made coffee for Tom and placed it on his desk so task is complete.
OK.
> done 
***
\end{verbatim}
\end{tcolorbox}

\subsection{VLM prompt structures}
\label{sec:app-vlm-prompt}

During experimentation, we identified an issue wherein the VLM exhibited hallucinations, falsely asserting that it could not interpret provided images. This behavior was particularly observed when the prompt included an additional instruction, such as specifying the output format. To mitigate this, we restructured the prompting strategy into a two-stage process: (1) the model generates a natural language description of the image, and (2) this description is subsequently parsed into a structured JSON format.

\textbf{VLM descriptor prompt:}

\begin{tcolorbox}[
  enhanced,                
  breakable,               
  colback=gray!10,      
  colframe=gray!50,     
  arc=4mm,              
  boxrule=0.8pt,        
  left=6pt, right=6pt,  
  top=6pt, bottom=6pt   
]
\noindent\textbf{List all objects ontop or inside of \texttt{*ASSET*}.} 
Specify the spatial relation (on top / inside) for each. If possible, use following names for objects: \texttt{*OBJ\_NAMES*}.\\
Previously, the following objects were spotted: \texttt{*List of prior objects related to asset*}. These objects can now be removed or new ones added — please proceed carefully.
\end{tcolorbox}

\textbf{VLM parser prompt:}

\begin{tcolorbox}[
  enhanced,                
  breakable,               
  colback=gray!10,      
  colframe=gray!50,     
  arc=4mm,              
  boxrule=0.8pt,        
  left=6pt, right=6pt,  
  top=6pt, bottom=6pt   
]
\small

Given the description of \texttt{ASSET} and the list of possible objects in the environment, build a subgraph of objects related only to \texttt{"ASSET"}. Answer in structured JSON format.\\

\noindent\textbf{Possible object names: \texttt{OBJ\_NAMES}}

\noindent\textbf{Return the results in a predefined JSON format as follows:}

\begin{verbatim}
[
  {
    "name": "object_name",
    "relation": "relation_type", 
    "related_to": "related_object_name",
    "states": "object_state"
  }
]
\end{verbatim}

\noindent\textbf{Example Output:}
\begin{verbatim}
[
  {
    "name": "dishbowl-1",
    "relation": ontop_of,
    "related_to": "bench-1",
    "states": ""
  },
  {
    "name": "apple-1",
    "relation": "inside_of",
    "related_to": "dishbowl-1",
    "states": ""
  },
  {
    "name": "apple-2",
    "relation": "inside_of",
    "related_to": "dishbowl-1",
    "states": ""
  },
  {
    "name": "bottle-1",
    "relation": null,
    "related_to": null,
    "states": "closed"
  }
]
\end{verbatim}

\noindent\textbf{Details:}\\
1. States: Possible states are: \texttt{open}, \texttt{closed}, \texttt{turned\_on}, \texttt{turned\_off}.\\
2. Relations: Possible relations are: \texttt{ontop\_of}, \texttt{inside\_of}. Use these exclusively to describe relations.\\
3. Null Values: If there is no state or relation, set the value as \texttt{null} or an empty string (\texttt{""}) as appropriate.\\
4. Don't assume anything.\\
5. Give every object only one name from possible object names.
\end{tcolorbox}

\subsection{Dataset preparation}
\label{app:datset_prep}
    
\subsubsection{VirtualHome Graph Construction}

The VirtualHome RobotHow dataset provides initial and goal graphs paired with natural language instructions. To convert these graphs into a hierarchical structure, we retain only the ``ON'' and ``INSIDE'' edges, removing all others. Each node is then classified using the properties and categories defined in the original graph.

To identify Place nodes, we use the ``Room'' category. To distinguish between assets and objects, we rely on the internal ``GRABBABLE'' property, which indicates whether an agent can pick up the item. Nodes that are not grabbable are classified as assets, while grabbable ones are considered objects. The state information for both objects and assets is directly copied from their internal states.

Additionally, we remove redundant nodes such as multiple floors, doors, ceilings, and walls, as these are not treated as objects in the 3DSG representation.

To focus only on tasks relevant to manipulation in the environment, we filter the dataset using both the ground truth (GT) plans and task names. We retain only those tasks that involve the following actions:

\textbf{Allowed Actions in VirtualHome:}
"FIND", "WALK", "GRAB", "SWITCHON", "TURNTO", "PUTBACK", "LOOKAT", "OPEN", "CLOSE", "PUTOBJBACK", "SWITCHOFF", "PUTIN", "RUN".

Furthermore, we exclude tasks that are irrelevant to robotic manipulation. The following task types are filtered out:

Open door, Lock door, Look out window, Movie, Clean, Write school paper, Dust, Play games, Get dressed, Playing video game, Shave, Print out papers, Watch fly, Walk through, Admire art, Gaze out window, Look at painting, Check appearance in mirror, Shut front door, Look at mirror, Write an email, Browse internet, Watch TV, Take shower, Work, Drink, Wash teeth, Wash dishes by hand, Pet cat, Brush teeth, Keep an eye on stove as something is cooking, Open front door, Close door, Pick up phone.

To reduce ambiguity, the task name was revised for greater specificity. For instance, the original label "Turn on light" was updated to "Turn on top light in *target room*."

\subsubsection{Omnigibson Graphs Construction}
The BEHAVIOR-1K dataset includes descriptions of 1,000 tasks relevant to real-world scenarios, paired with a OmniGibson simulator providing rooms, scenes, and PDDL task descriptions. We construct graph representations of the environment from the PDDL descriptions.
The initial graph is generated using a rule-based method based on the ontop and inroom predicates. Human annotators label the goal state, object relations, affordances, and states for each object in both the initial and goal graphs. 
Additionally, natural language task descriptions were created to accompany each example. For instance, the task ``assembling\_gift\_baskets-0'' is described as: ``Place one candle, one butter cookie, one piece of Swiss cheese, and one bow into each of the four wicker baskets, ensuring each basket contains one of each item type.''

To focus on manipulation tasks, we filter the dataset to exclude tasks requiring cooking or cleaning skills, and select only tasks with the predicates \texttt{ontop}, \texttt{real}, \texttt{inside}, \texttt{open}, and \texttt{toggled on}. This results in 177 tasks with initial and goal graph pairs. The graphs are constructed from task descriptions only, excluding unrelated nodes. This allows for a clear evaluation of planning performance without the need to filter out irrelevant nodes.

\subsubsection{SayPlan Office Graphs Construction}

We used the SayPlan Office dataset to evaluate our method on a variety of human-formulated tasks. Since the original dataset is not available, we reconstructed the environment graph representations based on details from the original paper. 
Since traversing through rooms via pose nodes could be addressed by the classical path planner from the original paper, we removed the pose nodes and directly connected the rooms to the scene nodes.
Graph representations and corresponding action sequences were constructed manually, selecting tasks from both simple and complex planning sections of the paper.
Using these reconstructed initial graphs and action sequences, we employed a Scene Graph Simulator to generate goal graph representations after task execution. Ambiguous tasks, such as ``Put an object into a place where I can enjoy it'' were paraphrased. The final dataset includes 29 curated tasks with initial and goal graph pairs.

\end{document}